\documentclass[letterpaper, 10 pt, conference]{ieeeconf}  

\IEEEoverridecommandlockouts                              

\usepackage[T1]{fontenc}
\usepackage{blindtext}
\usepackage{graphicx}
\usepackage[table]{xcolor}

\usepackage{amsmath}
\usepackage{amssymb}

\usepackage{graphicx}
\usepackage{placeins}

\usepackage{diagbox} 
\usepackage{subfig}
\usepackage{gensymb}
\usepackage{textcomp}
\usepackage{hyperref}
\usepackage{amsmath}

\usepackage{multirow}
\usepackage{balance}

\usepackage{amsmath}

\usepackage{mathtools}

\usepackage[colorinlistoftodos]{todonotes}

\usepackage{tabularx}
\usepackage{lipsum}

\usepackage{csvsimple}

\usepackage[free-standing-units=true]{siunitx}
\usepackage{booktabs}

\newcommand{\etalcite}[2]{#1~et~al.~\cite{#2}}

\usepackage{amssymb} 
\usepackage{siunitx}

\title{\LARGE \bf
Learning an Efficient Terrain Representation \\ for Haptic Localization of a Legged Robot
} 

\author{Damian S\'{o}jka \and Micha\l{} R. Nowicki \and Piotr Skrzypczy\'nski
\thanks{Institute of Robotics and Machine Intelligence, Poznan University of Technology, Poznan, Poland
        {\tt\small michal.nowicki@put.poznan.pl}}
\thanks{
*M. R. Nowicki is supported by the Foundation for Polish Science (FNP).
}
}

\begin{document}

\maketitle
\thispagestyle{empty}
\pagestyle{empty}

\begin{abstract}
Although haptic sensing has recently been used for legged robot localization in extreme environments where a camera or LiDAR might fail, the problem of efficiently representing the haptic signatures in a learned prior map is still open. 
This paper introduces an approach to terrain representation for haptic localization inspired by recent trends in machine learning. It combines this approach with the proven Monte Carlo algorithm to obtain an accurate, computation-efficient, and practical method for localizing legged robots under adversarial environmental conditions. 
We apply the triplet loss concept to learn highly descriptive embeddings in a transformer-based neural network.
As the training haptic data are not labeled, the positive and negative examples are discriminated by their geometric locations discovered while training. 
We demonstrate experimentally that the proposed approach outperforms by a large margin the previous solutions to haptic localization of legged robots concerning the accuracy, inference time, and the amount of data stored in the map. 
As far as we know, this is the first approach that completely removes the need to
use a dense terrain map for accurate haptic localization, thus paving the way to practical applications.
\end{abstract}

\section{Introduction}
Recent years have brought legged locomotion from labs to real-world applications, focusing on inspection or search-and-rescue tasks in harsh environments like industrial facilities, disaster sites, or mines \cite{zimroz2019mining}.
So far, few works have demonstrated the possibility of localizing a walking robot without visual or LiDAR-based SLAM, employing haptic sensing, using signals from IMUs, force/torque (F/T) sensors in the feet, and joint encoders \cite{chitta2007icra,buchanan2020haptic,hapticar}.
Whereas these papers demonstrated a possibility of solving the pose tracking problem employing the Monte Carlo Localization (MCL) algorithm with particle filtering, the representation of the terrain and foot/terrain interactions extracted from haptic information remained an open problem. 
This representation is essential for haptic localization, as interactions between the robot's feet and the terrain are the only source of exteroceptive information in this problem formulation. 
Hence, haptic information representation must be descriptive enough to distinguish between the steps taken at different locations, even if these footholds are located on a similar surface. 
Moreover, this representation needs to be compact to allow quick retrieval of the data from terrain map and efficient comparison of the locations.
The practical aspect of the representation problem is how the terrain map is obtained. 
A dense 2.5D elevation map used in \cite{buchanan2020haptic} has to be surveyed using an external LiDAR sensor. In contrast, a map of terrain types encoded as classes on a grid map \cite{hapticar} needs tedious manual labeling.
Both approaches confine the operation of a walking robot to small-scale pre-surveyed environments making the haptic localization concept rather impractical in real-world applications.

\begin{figure}[t!]
    \centering
    \includegraphics[trim={0 0 0.5cm 0},clip,width=\columnwidth]{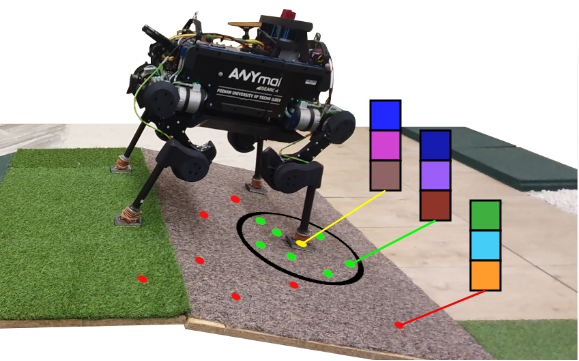}
    \caption{Haptic localization requires a distinctive representation of the foot/terrain interaction to distinguish between locations. We propose to train a transformer-based neural network with triplet loss to minimize the difference between embeddings for steps close to each other while maximizing this difference for steps further away.}
    \label{fig:catchy}
    \vspace{-5mm}
\end{figure}

In contrast, this research uses the building blocks of machine learning methods already proven in computer vision \cite{ho2020triplet} and place recognition \cite{yu2020netvlad} problems to create a sparse map of highly descriptive signatures in the locations touched by the robot's feet.
This concept leverages the possibility of training a neural network with triplet loss to extract the features that differentiate the neighboring footholds from the collected haptic signals while suppressing those irrelevant for localization (Fig.~\ref{fig:catchy}). 
Interestingly, this approach addresses both the mentioned challenges, creating embeddings (latent vectors of the signatures) that are simultaneously highly descriptive and extremely compact. 
Our approach uses a new neural network architecture based on parameter-efficient transformer layers to build embeddings for a sparse terrain map, which ensures short inference times to achieve real-time operation of the haptic MCL. 
The contribution of our work can be summarized as follows:
\begin{itemize}
    \item The first adaptation of the triplet loss training paradigm for learning local terrain representations from haptic information that lacks  
          explicit class labels for the positive and negative examples.
    \item An efficient transformer-based neural network architecture for computing the embeddings.
    \item A novel variant of the MCL method for legged robots that employs a sparse map of haptic embeddings and 3D positions of steps. It allows the robot to 
          self-localize using only haptic information without any map that needs to be created or annotated manually. 
\end{itemize}

\section{Related work}

Walking robots commonly use haptic information from their legs in terrain classification and gait adaptation. 
The approaches to terrain classification~\cite{Hoepflinger2010,Wu2016,bednarek19,kolvenbach19} demonstrated that haptic information, like IMU or force/torque signals, can be used to determine the class of the terrain the robot is walking on. 
These methods achieve similar, high accuracies~\cite{bednarek2019icra} but the supervised approaches to terrain classification are data-hungry and offer poor generalization to unseen classes.
Moreover, the choice of these classes needs too be known upfront and is based on human perception.
To overcome these challenges, \etalcite{Bednarek}{ecmr} proposed an efficient transformer-based model resulting in fast inference and a decreased need for samples to train the network while improving the robustness of the solution to noisy or previously unseen data. 
Another attempt to decrease the required number of samples is~\cite{Ahmadi2021}, which uses a semi-supervised approach to the training of a Recurrent Neural Network (RNN) based on Gated Recurrent Units. 

Gait adaptation is commonly mentioned as one of the applications of terrain classification. 
\etalcite{Lee}{hutter2020science} proved that end-to-end learning can generate walking policies that adapt well to the changing environment. In this approach, driven by simulation, the terrain information is represented internally without explicit classes, while more recently \cite{miki2022} demonstrated a learned robust controller that combines terrain map data and proprioceptive locomotion when needed.
\etalcite{Gangapurwala}{oxfordtro22} employ reinforcement learning policies to prioritize stability over aggressive locomotion while achieving the desired whole-body motion tracking and recovery control in legged locomotion. 
Despite progress with end-to-end approaches, we also see works that benefit from combining the trained and classical model-based approaches.
One example is the work of \etalcite{Yuntao}{Yuntao22}, who shows that model-based predictive control can predict future interactions to increase the robustness of training policies. 

Our haptic localization system takes inspiration from both of the presented domains. It follows the general processing scheme introduced by \etalcite{Buchanan}{buchanan2020haptic}, who proposed an MCL algorithm for walking robots based on the measured terrain height and a dense 2.5D map of this terrain built with an accurate external 3D LiDAR.
Their approach uses the relative height of the leg touching the ground to compute the updated particle positions in MCL, thus reducing the accumulation of the localization drift. 
Moreover, their follow-up work~\cite{hapticar} introduced a haptic localization system that additionally utilizes the terrain classification information to further improve the localization accuracy, as long as the dense 2.5D map has prior class labels for each cell of the map.
While this work demonstrated that geometric information is complementary to tactile sensing, it also revealed the limitations of terrain classification employing a discrete number of terrain classes that might not have strict borders when applied in a real-world scenario.
In this context, \etalcite{Łysakowski}{lysakowski2022unsupervised} showed that localization could be performed with compressed tactile information using Improved AutoEncoders, thus avoiding explicit terrain classification.
Moreover, their approach demonstrated the possibility of working with a sparse map of latent signal representations, making it possible to learn the terrain map by the robot itself without tedious manual labeling. 
But the terrain representation from \cite{lysakowski2022unsupervised} just compresses the haptic signals, without selecting valuable features.

In this work, we propose a new HL-ST approach to haptic localization 
using a sparse geometric map and a latent representation of the haptic information that benefits from a training scheme with triplet loss, which has not yet been used in training on haptic signals for localization problems.
This stands in contrast to \cite{lysakowski2022unsupervised}, where the training process is fully unsupervised, thus giving no control over the learned representation. 
Similarly to~\cite{ecmr}, we also employ a transformer-based architecture to achieve a parameter-efficient network, but we train it to generate embeddings rather than class labels.
Moreover, inspired by works in gait adaptation~\cite{hutter2020science}, the critical ingredient of our localization solution (latent representation/embedding) is trained to benefit from a large number of collected samples.

\section{Proposed Method}

Our problem statement is driven by an application of a walking robot performing repetitive tasks over a known route. 
We would like to quickly explore the desired path with the robot and then operate solely based on the legged odometry and haptic signals, even in challenging environments.
In contrast to~\cite{buchanan2020haptic, hapticar}, we assume no prior dense map, only requiring accurate localization for the first walk along the given route. 

\begin{figure}[htbp!]
    \centering
    \includegraphics[width=\columnwidth]{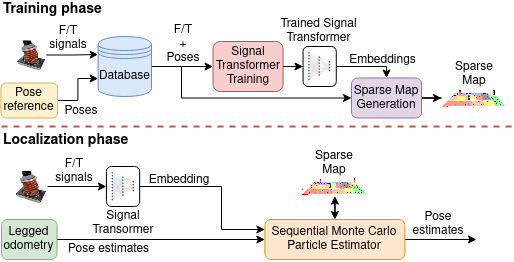}
    \caption{Overview of the training and processing pipelines for our Haptic Localization utilizing Signal Transformer.}
    \label{fig:system}
    \vspace{-3mm}
\end{figure}

\begin{figure*}[htbp!]
    \centering
    \vspace{3mm}
    \includegraphics[width=\textwidth]{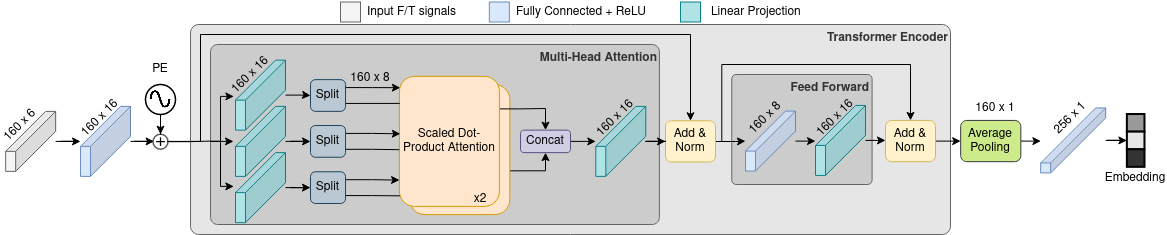}
    \caption{The proposed Signal Transformer network that processes time sequence of force/torque signals to generate a location descriptive embedding.}
    \label{fig:network}
    \vspace{-5mm}
\end{figure*}

An overview of our approach is presented in Fig.~\ref{fig:system}. 
Each localization event is triggered by a foot placement on the ground that captures 160 consecutive samples from the F/T sensors mounted at that foot.
In the initial phase, each step event has an associated localization estimate (e.g., from SLAM), and the whole sequence is used to gather a database of signals. 
This database is used to train our transformer-based network on triplets of samples to find the latent representation that best suits the localization purposes.
The trained network processes the entire database to create a sparse map of embeddings at the measured locations.

During the localization phase, raw measurements from step events are fed to the trained neural network to obtain embeddings that can be compared to those already stored in the sparse terrain map.
These comparisons are used to update our localization estimates represented by particles in the MCL framework.

\subsection{Learning Terrain Representation}
\label{sec:unsupervised-measurement}

The critical component and our main contribution is the network to determine the embeddings that encode relevant features from the raw haptic input. The network is called Signal Transformer, based on the original transformer architecture from \cite{vaswani2017attention} and is presented in Fig.~\ref{fig:network}.
The 6-dimensional sensor input (from 3-axis force and 3-axis torque sensors) for 160 consecutive measurements is converted into a 16-dimensional feature space with a fully-connected layer. We apply layer normalization and augment the sequence with learnable positional encoding. Augmented data are passed to the encoder of the reference transformer architecture from \cite{vaswani2017attention}, with $h=2$ attention heads, the dimensionality of model $d_{m}=16$, and the size of inner feed-forward layer $d_{ff}=8$. The use of average pooling flattens output from the encoder. The final latent representation is generated by applying batch normalization and feeding normalized data to the dense feed-forward layer with the ReLU activation function. 
The final layer has the number of neurons equal to the length of the embedding, which by default is set to 256, as in \cite{lysakowski2022unsupervised}. Implementation of the proposed network is publicly available\footnote{https://github.com/dmn-sjk/signal\_transformer}.

\begin{figure}[h!]
    \centering
    \includegraphics[width=\columnwidth]{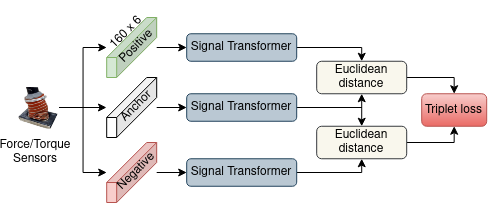}
    \caption{The Signal Transformer network is trained using a triplet of data samples to achieve the desired similarity of embeddings for the anchor and positive sample while increasing the difference for the anchor and negative sample.}
    \label{fig:triplet}
    \vspace{-5mm}
\end{figure}

The network is trained with triplet loss presented in Fig.~\ref{fig:triplet}. For efficiency, an online triplet mining technique is used to reduce the number of inactive triplets and to improve the convergence and speed of training.
We form mini-batches randomly, with every example considered an anchor during the training. 
We need an associated location where this sample was captured for each training sample. Positive examples for a specific anchor are those sampled closer to the anchor than the defined constant distance threshold $d_{\rm thr}$. 
Consequently, the negative examples were sampled further away than $d_{\rm thr}$:
\begin{equation}
    \left\{
    \begin{array}{cc}
        d(\mathbf{s}_{b_a}, \mathbf{s}_{b_i}) > d_{\rm thr} \rightarrow b_i \in N_a \\
        d(\mathbf{s}_{b_a}, \mathbf{s}_{b_i}) \leq d_{\rm thr} \rightarrow b_i \in P_a,
    \end{array}
    \right.
\end{equation}
where $d(\mathbf{s}_{b_a}, \mathbf{s}_{b_i})$ is the Euclidean distance between step position of an anchor $\mathbf{s}_{b_a}$ and step position $\mathbf{s}_{b_i}$ of the $i$-th data sample $b_i$. $P_a$ and $N_a$ denote a set of positives and negatives concerning the $a$-th anchor.
The distance threshold $d_{\rm thr}$ is a hyperparameter that can be adjusted. We used $d_{\rm thr}=25$ cm since it provided the best localization accuracy.
During training, positive and negative examples depend strictly on the spatial dependencies between step positions without any terrain class labels.
Inspired by \cite{hermans2017defense}, we use Batch All triplet loss variation, but without special mini-batch sampling, considering the lack of class annotations, and calculate it as:
\begin{equation}
\label{eq:tlba}
    \mathcal{L} = \sum_{a=1}^B \sum_{p=1}^{|P_a|} \sum_{n=1}^{|N_a|} \left[d(f(b_a), f(b_p)) - d(f(b_a), f(b_n)) + m \right]_+,
\end{equation}
where $B$ is the size of the mini-batch, $m$ indicates the margin, and $|\cdot|$ denotes the cardinality of sets.
The distance function $d(\cdot)$ is implementing an Euclidean distance. 
The average of Batch All triplet loss is calculated considering only these triplets, which have a non-zero loss as in~\cite{hermans2017defense}.
The training process was performed using AdamW optimizer from \cite{adamw}. The learning rate was exponentially decreased with an initial value of \num{5e-4}. The initial value of weight decay was equal to \num{2e-4} and was reduced with cosine decay. Mini-batch size was set to 128. The training lasted for 200 epochs.

\subsection{Sparse Haptic Map Generation}
\label{sec:sparsemap}

\begin{figure}[h!]
    \centering
    \includegraphics[width=\columnwidth]{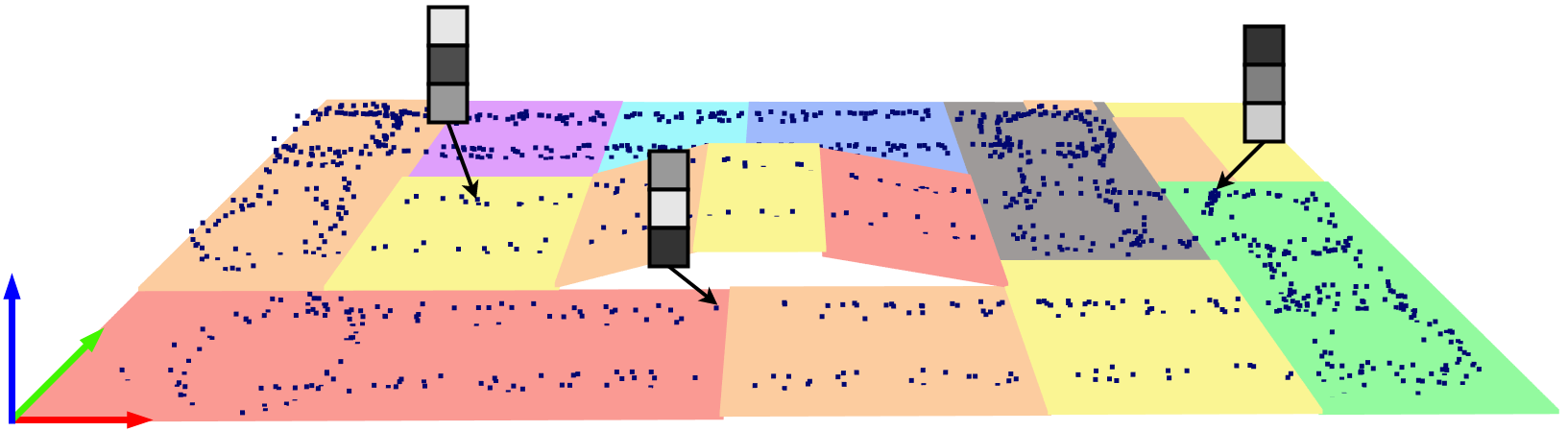}
    \caption{Sparse haptic map visualization. Dark blue points indicate footholds recorded during the mapping run. Each 2D foothold holds an embedding and terrain elevation value. Color patches distinguish terrain classes in the PUTany dataset, but class labels are not used in our new approach.}
    \label{fig:sparse_map}
    \vspace{-5mm}
\end{figure}

The proposed network is used to build a sparse haptic map of embedding vectors visualized in Fig.~\ref{fig:sparse_map}. 
During the initial run, raw measurements taken at the contact of each foot with the ground are recorded along with the reference robot's position.
The network is then trained with the triplet loss using these data. 
Once training is completed, data from each step are passed through the network to obtain the reduced latent representation (embedding), which is added to the map at the exact location of this step's foothold. 
The resulting map is sparse and unevenly distributed.
The map generation phase requires an independent source of 6 DoF robot's pose estimates to properly train the network (generation of positive and negative examples) and to place the inferred embeddings in the map accurately.
In practical applications, an onboard LiDAR-based localization subsystem of the robot \cite{ramezani2020slam} can be utilized as this independent source.

\subsection{Sequential Monte Carlo Localization}
\label{sec:primer}
The proposed neural network generates highly descriptive embeddings -- a latent representation of tactile sensing. 
To verify its impact on the localization, we use the sequential MCL algorithm proposed in~\cite{hapticar} that was also used in the follow-up works on unsupervised terrain localization~\cite{lysakowski2022unsupervised}. 
We present the general idea of this method while noting that this part of the processing is not a contribution of this paper.

Given a history of measurements $\boldsymbol{z}_0,\dots,\boldsymbol{z}_k = \boldsymbol{z}_{0:k}$, the MCL algorithm estimates the most likely pose $\boldsymbol{x}^{*}_k \in SE(3)$ at time $k$:
\begin{equation}
 p\left(\boldsymbol{x}_k | \boldsymbol{z}_{0:k}\right) = \sum_i w^i_{k-1}
 p\left(\boldsymbol{z}_k |
\boldsymbol{x}_k\right)p\left(\boldsymbol{x}_k|\boldsymbol{x}^i_{k-1}\right)
 \end{equation}
where $w^i$ is the \emph{importance weight} of the $i$-th particle, $p\left(\boldsymbol{z}_k,
\boldsymbol{x}_k\right)$ is the measurement likelihood function, and $p\left(\boldsymbol{x}_k|\boldsymbol{x}^i_{k-1}\right)$ is the motion model for the $i$-th particle state.

The system gets the measurements whenever the robot's foot touches the ground while using a statically stable gait.
Once measurements are taken, the probability of the particles is updated, particles are resampled, and a new pose estimate is available.

\subsection{Haptic Measurement Model}
\label{sec:localization-haptic}

For each taken step, the proposed neural network generates an embedding $\mathbf{v}$, based on signals from a single foot $f$ placed at the ground at the location $\mathbf{d}_f$, in the base coordinates:
\begin{equation}
\mathbf{d}^i_f = (d^i_{xf}, d^i_{yf}, d^i_{zf}) = \boldsymbol{x}^i_k \mathbf{d}_f.
\end{equation}

The position of the foot $\mathbf{d}^i_f$ is used to retrieve the haptic map entry $\mathbf{m} = \mathcal{S}(s^i_{xf},s^i_{yf})$ located in the map at $(s^i_{xf}, s^i_{yf})$, which is the nearest neighbor in 2D. 
For the sake of speed, search is based on 2D Euclidean distance with a map matching error $d_{2D} = \sqrt{\left(d^i_{xf} - s^i_{xf}\right)^2, \left(d^i_{yf} - s^i_{yf}\right)^2}$. 
The map entry $\mathbf{m}$ contains both the embedding of the haptic signal $\mathbf{w}$ and the elevation of the original measurement $e$.



Although we search for the map entries in 2D, we compare 3D positions for the purpose of the MCL measurement model taking advantage of the elevation values stored in the sparse map. 
We compare the latent representations using the $L_2$ norm:
\begin{equation}
    d_l = f_{L_2}({\bf v},{\bf w}) = \sqrt{\sum_{j=1}^{n} \left(v_j - w_j\right)^2},
\end{equation}
where $v_j, w_j$ encoding $j$-th component of the embeddings and $d_l$ is the latent representation distance. To include the elevation information, $d_e$ is defined as the difference between $z$-axis component of estimated foot position $d_{zf}^i$ and the elevation $e$ saved in the closest map entry:
\begin{equation}
    d_e = d_{zf}^i - e.
\end{equation}
We use univariate Gaussian distribution centered at the cell matched in the sparse latent map:
\begin{align}
p(z_k | \boldsymbol{x}^i_k)  = \begin{cases}
p_{min} & d_{2D} > d_{t}\\
\mathcal{N}(d_l,\sigma_l)  \mathcal{N}(d_{2D},\sigma_{2D}) \mathcal{N}(d_e,\sigma_e) & d_{2D} \leq d_{t},
\end{cases}
\label{eq:terrainlikelihood}
\end{align}
where $d_{t}=25$ cm is the Euclidean threshold when a step is considered to have a proper match in the sparse map, otherwise the probability value is set to $p_{min}=0.001$. The impact of haptic, 2D geometric component, and elevation is weighted by the experimentally determined sigma values $\sigma_l=0.4$, $\sigma_{2D}=0.4$, and $\sigma_{e}=0.01$, respectively.


We denote our haptic localization method as HL-ST when all source of information are read from a sparse map, HL-T when elevation is not considered, and HL-GT when elevation comes from a dense geometric map and is used in a separate measurement model 
as in~\cite{hapticar}.


\section{Experimental Results}

\subsection{Dataset and Ground Truth}

In the presented evaluation, we use the PUTany dataset proposed in~\cite{hapticar} that was already applied to evaluate other haptic localization systems.
The dataset consists of three trials (walks) of a quadruped robot ANYmal B300 robot equipped with F/T sensors in the feet conducted over a $3.5 \times 7$ m area containing uneven terrain with eight different terrain types ranging from ceramic to sand. 
The total distance traveled by the robot equals 715 m, with 6658 steps and a duration of 4054 s. 
We used a different sequence gathered on the same route to train the neural network proposed in this work.
The ground truth for the robot motion is captured with the millimeter-accuracy motion capture system (OptiTrack), providing 6 DoF robot poses at 100~Hz. 

\subsection{Accuracy measures for haptic localization}

The robot 
generates a trajectory of 6 DoF poses that can be compared to the ground truth trajectory to determine the accuracy of the localization method. 
We use the Absolute Pose Error ($APE$) metric that is computed for a single relative pose ${\bf T}$ between the estimated pose ${\bf P}$ and the ground truth pose ${\bf G}$ at the time stamp $t$ \cite{grupp2017evo}:
\begin{equation}
    {\bf T} = {\bf P}_{t}^{-1} {\bf G}_{t},
\end{equation}
where ${\bf P}_{t}, {\bf G}_{t} \in SE(3)$ are poses either available at time stamp $t$ or interpolated for the selected time stamp. 
In our evaluation, we follow the error metrics reported in the previous articles concerning haptic localization~\cite{buchanan2020haptic,hapticar,lysakowski2022unsupervised} using the 3D translational part of the error ${\bf T}$ calling it ${\bf t}_{3D}$.
The accuracy of our method is compared to the previously published results taking these results directly from the respective papers \cite{hapticar,lysakowski2022unsupervised}.

As already observed in \cite{lysakowski2022unsupervised}, the earlier haptic terrain recognition methods are not constraining the localization of particles in MCL in the vertical direction (parallel to the gravity vector), because the latent information is stored in a 2D array and lacks an elevation component. 
Therefore, in our comparisons, we also include the 2D translation error ${\bf t}_{2D}$ that ignores the error in the elevation. 
The ${\bf t}_{2D}$ results for the state-of-the-art methods were computed based on the publicly available source code of these systems.

\subsection{Haptic localization with latent map (without geometry)}
\label{sec:exp-latent-only}

Using a dense and accurate terrain map of the environment for robot localization is impractical, as creating such a map usually requires deploying a survey-grade LiDAR in the target environment.  
Therefore, we first consider a scenario when only haptic signals and accurate localization are available for the robot's training run, with the following runs relying solely on haptics for localization. 
With these conditions, neither state-of-the-art methods for haptic localization (HL) can use elevation information. 
As other methods cannot use an elevation map, we also constrain our HL-T system to ignore the elevation data stored in the sparse map, using only haptic data. 
We compare our work to HL-C~\cite{hapticar} utilizing terrain classification and HL-U~\cite{lysakowski2022unsupervised}, which uses unsupervised haptic latent representation.

All these methods are evaluated using the ${\bf t}_{2D}$ error as the elevation component is unconstrained, following the odometry drift. 
The obtained results are presented in Tab.~\ref{tab:final_haptic}.

\begin{table}[h!]
\centering

      \begin{tabular}{l|cccc} \toprule
         &  \textbf{TSIF}~\cite{bloesch2017ral} & \textbf{HL-C}~\cite{hapticar} & \textbf{HL-U}~\cite{lysakowski2022unsupervised} & \textbf{HL-T} \\ 
        \textbf{Trial} & ${\bf t}_{2D}$ & ${\bf t}_{2D}$ & ${\bf t}_{2D}$ & ${\bf t}_{2D}$  \\
        \midrule
        1 & 0.34 & 0.39 &  0.17 &  \textbf{0.07} \\
        2 & 0.92 & 0.22 &  0.14 &  \textbf{0.06} \\
        3 & 0.51 & 0.29 &  0.18 &  \textbf{0.08} \\
        \bottomrule
    \end{tabular}
\caption{Comparison of the 2D  Absolute Pose Error (APE, in [m]) for localization with haptic sensing only. The new HL-T method achieved the lowest error on all sequences.}
\label{tab:final_haptic}
\vspace{-3mm}
\end{table}

The worst results are produced by TSIF, the legged odometry estimator, which is both a baseline solution and a component of the MCL method in the remaining systems. 
Among the compared solutions, the proposed HL-T outperforms previous approaches by a large margin, reducing the APE values by almost 50\%. 
We believe it stems from the fact that our method is not constrained to a limited number of discrete classes like HL-C, while unlike HL-U,
it can learn an internal representation of the haptic signals that promote discriminative features.
What is important, the error did not exceed 10 cm despite the lack of other sensing modalities, which may be sufficient to let an autonomous robot continue its operation despite a vision-based sensor failure or a sudden change in the environmental conditions.



\FloatBarrier

\subsection{Haptic localization with dense geometric and sparse latent map}
\label{sec:res_loca}

Let's consider scenarios when an accurate 2.5D map of the environment is available for localization purposes.
For such scenarios, we have HL-G~\cite{buchanan2020haptic} utilizing the dense height map of the terrain for pose correction, HL-GC~\cite{hapticar} utilizing both the geometry and terrain classification, and HL-GU~\cite{lysakowski2022unsupervised} which uses the geometry and unsupervised haptic latent representation. 
In these experiments, our HL-GT method is configured to use a dense elevation map and a sparse latent map.
The results of the legged odometry estimator TSIF~\cite{bloesch2017ral} were omitted as it has already been proven that HL-G, HL-GC, and HL-GU outperform it in these trials.
The results for both types of errors (${\bf t}_{2D}$, ${\bf t}_{3D}$) are presented in Tab.~\ref{tab:haptic}.

\begin{table}[htbp!]
\vspace{-1mm}
\centering
    \begin{tabular}{l|cccccccc} \toprule
         &  \multicolumn{2}{c}{\textbf{HL-G}~\cite{hapticar}}  & \multicolumn{2}{c}{\textbf{HL-GC}~\cite{hapticar}} & \multicolumn{2}{c}{\textbf{HL-GU}~\cite{lysakowski2022unsupervised}} & \multicolumn{2}{c}{\textbf{HL-GT}} \\ 
        \textbf{Trial} &  ${\bf t}_{3D}$ & ${\bf t}_{2D}$ & ${\bf t}_{3D}$ & ${\bf t}_{2D}$ & ${\bf t}_{3D}$ & ${\bf t}_{2D}$ & ${\bf t}_{3D}$ & ${\bf t}_{2D}$ \\
        \midrule
        1 & 0.23 & 0.23 &  0.14 & 0.12 & 0.15 & 0.09 & \textbf{0.09} & \textbf{0.08} \\
        2 & 0.25 & 0.20 & \textbf{0.11} & 0.11 & 0.18 & 0.12 & \textbf{0.11} &  \textbf{0.10} \\
        3 & 0.21 & 0.18 & 0.18 & 0.17 & 0.13 & 0.13 & \textbf{0.09} & \textbf{0.09} \\
        \bottomrule
    \end{tabular}
\caption{Comparison of the 3D and 2D Absolute Pose Error (APE, in [m]) for localization solutions utilizing both prior dense geometric map and haptic terrain recognition solutions. Proposed HL-GT provides the best results using both ${\bf t}_{3D}$ and ${\bf t}_{2D}$ error metric.}
\label{tab:haptic}
\vspace{-2mm}
\end{table}

The obtained results for all considered methods show that the ${\bf t}_{3D}$ and ${\bf t}_{2D}$ errors almost match each other, proving that there is no significant drift in the elevation direction due to the availability of the dense elevation map. 
The proposed HL-GT outperforms other solutions, which suggests that the latent representation trained with triplet loss is more suited for 
distinguishing between terrain locations than terrain classification (HL-GC) or unsupervised terrain representation/signal compression (HL-GU).
Moreover, the haptic signal information encoding in HL-GT is complementary to the dense elevation map as the method improves the performance of the bare geometric approach (HL-G).

\FloatBarrier

\subsection{Haptic localization with sparse geometric map}

One of the advantages of the proposed solution is the ability to use elevation information even if only localization ground truth was available for training. 
This improvement significantly impacts the solution's practicality as no survey-grade LiDAR is required to take advantage of the elevation data.
Therefore, we decided to compare three solutions: HL-T, which uses solely haptic signals for localization, HL-GT which uses dense geometric map and haptic signals, and HL-ST, which uses sparse geometric and latent map.
The results are presented in Tab.~\ref{tab:our}.

\begin{table}[htbp!]
\vspace{-1mm}
\centering
    \begin{tabular}{l|cccccc} \toprule
         &  \multicolumn{2}{c}{\textbf{HL-T}}  & \multicolumn{2}{c}{\textbf{HL-GT}} & \multicolumn{2}{c}{\textbf{HL-ST}} \\ 
        \textbf{Trial} &  ${\bf t}_{3D}$ & ${\bf t}_{2D}$ & ${\bf t}_{3D}$ & ${\bf t}_{2D}$ & ${\bf t}_{3D}$ & ${\bf t}_{2D}$ \\
        \midrule
        1 & 0.51 & \textbf{0.07} & \textbf{0.09} & 0.08 & \textbf{0.09} & 0.09 \\
        2 & 0.77 & \textbf{0.06} & \textbf{0.11} & 0.10 & \textbf{0.11} & 0.11 \\
        3 & 0.44 & \textbf{0.08} & \textbf{0.09} & 0.09 & 0.10 & 0.09 \\
        \bottomrule
    \end{tabular}
\caption{Comparison of the 3D and 2D Absolute Pose Error (APE, in [m]) for localization without geometry (HL-T), with dense geometric map (HL-GT), and sparse geometric map (HL-ST). HL-ST performs similarly to HL-GT in 2D and 3D without a tedious mapping phase.}
\label{tab:our}
\vspace{-2mm}
\end{table}

The results show that the HL-T approach provides the best results in 2D. Still, the 3D error is unbounded following the legged odometry's general drift, making it impractical for any autonomous operation. 
On the other hand, HL-GT provides the most accurate 3D localization due to the dense geometric map.
The proposed HL-ST is a good trade-off between these approaches as the 2D and 3D errors are comparable with HL-T and HL-GT while only using the haptics and localization for the first trial.
We believe HL-ST is, therefore, a unique solution that may support legged robot autonomy in challenging, real-world applications.

\subsection{Inference time evaluation}
\label{sec:inf_time_exp}

Autonomous operation requires real-time processing with short inference times.
We compared the average inference times of the proposed Signal Transformer network with the classification neural network from HL-C~\cite{hapticar} and the unsupervised network HL-U~\cite{lysakowski2022unsupervised}. Inference times were measured on over 10 000 samples on NVIDIA GeForce GTX 1050 Mobile GPU matching a similarly capable GPU that can fit in a walking robot. 
Table \ref{tab:inference_time_results} contains the obtained results.

\begin{table}[htbp!]
\vspace{-1mm}
\centering
    \begin{tabular}{c|ccc} \toprule
        HL method & HL-C & HL-U & HL-T \\ 
        \midrule
        Inference time [ms] & 21.20 $\pm$ 2.94 & 30.72 $\pm$ 4.84 & 2.20 $\pm$ 0.28 \\
        \bottomrule
    \end{tabular}
\caption{Inference time comparison between models used to process the haptic signal for localization purposes.}
\label{tab:inference_time_results}
\vspace{-2mm}
\end{table}

The inference time of the Signal Transformer being a part of the HL-T/HL-ST solutions is one magnitude smaller when compared to the inference times of neural networks used in HL-C and HL-U. 
The observed gains stem from a reduced number of parameters for our networks (45992 parameters) compared to over 1 million parameters for networks used in HL-C and HL-U. The transformer-based architecture proved to be more compact and suitable for on-board deployment in a robot than previously used solutions. 

\subsection{The size of the latent representation}
\label{sec:lv_sizes_exp}

\begin{figure}[h!]
    \centering
    \includegraphics[width=\columnwidth]{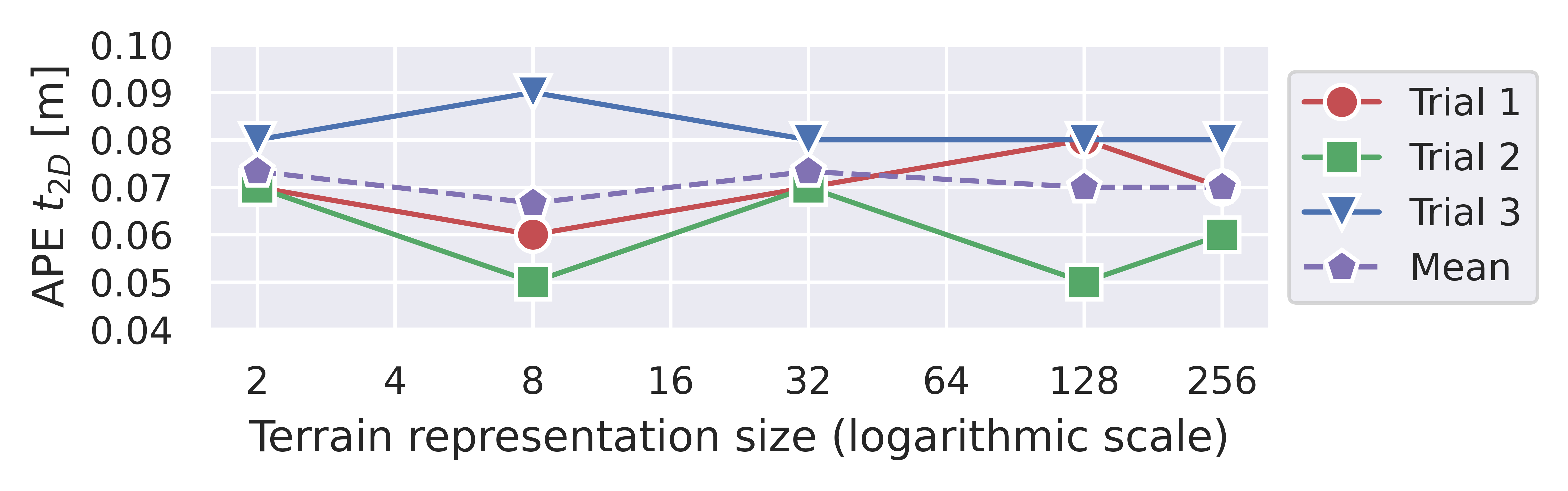}
    \caption{The graph of the HL-T method's 2D localization error ${\bf t}_{2D}$ as a function of the trained embedding size for the proposed Signal Transformer architecture. The Mean is the average outcome of all three trials. We see no major difference in APE, even with a small embedding size.}
    \label{fig:latentsize}
    \vspace{-4mm}
\end{figure}

The transformer-based architectures are known to be efficient, needing almost a fraction of the resources (parameters and inference time) of other known architectures to achieve comparable results. 
Moreover, learning with triplet loss can train a representation with desired characteristics. 
Therefore, we wanted to verify the required size of the embeddings necessary to achieve good localization results using a more challenging HL-T approach.
The obtained APEs depending on the chosen size of the embeddings are presented in Fig.~\ref{fig:latentsize}.

Decreasing the embedding size from the original size of 256 did not affect the localization accuracy,
suggesting that an embedding vector with a length as low as 2 contains enough information to distinguish between embeddings from multiple positions in a given environment. 
This result is of practical importance, as for a possible map containing 10 000 steps, the original size of 75 MB for embeddings with length 256 can be reduced to 1.4 MB using embeddings of size 2, which means a substantial reduction in the amount of stored data, a possibility to operate over a larger area, and the reduction of map matching times.

\section{Conclusions}
This work investigated how to employ triplet loss to train a Signal Transformer network to compute descriptive embeddings from haptic signals. 
The experiments indicate that the novel localization method employing these embeddings outperforms state-of-the-art haptic localization solutions (HL-C and HL-U) when only haptics are used in a controlled environment. At the same time, the HL-GT variant achieves the lowest localization error (3D APE, compared to HL-GC and HL-GU) when a dense 2.5D map is used due to an efficient representation of the haptic data.
In contrast to previous works, we can build and use a sparse geometric map (HL-ST), resulting in a practical solution requiring only reference poses for the first robot run while achieving results on par with solutions utilizing dense 2.5D maps.
Moreover, we show that our network can process data 10 times faster than previous approaches and can reduce the embeddings from a size of 256 to 2 without a significant impact on the localization error.
As a result, we achieve a haptic localization method that is more practical than state-of-the-art solutions. 
Our future work will concern outdoor experiments in scenarios without clear terrain type boundaries.

\balance

\bibliographystyle{IEEEtran}
\bibliography{biblio}

\end{document}